\pdfoutput=1

\documentclass[11pt]{article}

\usepackage[review]{ACL2023}

\usepackage{times}
\usepackage{latexsym}
\usepackage{graphicx,xcolor}
\usepackage{pxrubrica}
\usepackage{url}
\usepackage{enumitem}
\usepackage{array}
\usepackage{booktabs}
\usepackage[colorinlistoftodos]{todonotes}
\usepackage{tabularray}
\usepackage{float}

\usepackage[T1]{fontenc}

\usepackage[utf8]{inputenc}

\usepackage{microtype}

\usepackage{inconsolata}

\usepackage{soul}
\usepackage{color}

\DeclareRobustCommand{\hlcolor}[2]{{\sethlcolor{#1}\hl{#2}}}

\newcommand*{\escape}[1]{\texttt{\textbackslash#1}}
\definecolor{blue-violet}{rgb}{0.54, 0.17, 0.89}
\definecolor{bleudefrance}{rgb}{0.19, 0.55, 0.91}
\definecolor{babyblueeyes}{rgb}{0.63, 0.79, 0.95}

\newcommand{\prevwcolor}{black!10}
\newcommand{\proposedwcolor}{babyblueeyes!50}

\title{Flee the Flaw: Annotating the Underlying Logic of\\Fallacious Arguments Through Templates and Slot-filling}

\author{%
  Irfan Robbani${}^{1}$ Paul Reisert${}^{2}$ Naoya Inoue${}^{1,3}$ Surawat Pothong${}^{1}$
    Camélia Guerraoui${}^{4,3,5}$\quad\\
    \textbf{Wenzhi Wang}${}^{4,3}$\quad
    \textbf{Shoichi Naito}${}^{6,4,3}$\quad
    \textbf{Jungmin Choi}${}^{3}$\quad
    \textbf{Kentaro Inui}${}^{7,4,3}$\\
${}^{1}$JAIST\quad
 ${}^{2}$Beyond Reason\quad
 ${}^{3}$RIKEN\quad
 ${}^{4}$Tohoku University\\ 
 ${}^{5}$INSA Lyon\quad
 ${}^{6}$Ricoh Company, Ltd.\quad
 ${}^{7}$MBZUAI\\     
 \small
 \texttt{\{robbaniirfan,naoya-i,spothong\}@jaist.ac.jp} \quad\texttt{beyond.reason.sp@gmail.com}\\
  \small
 \texttt{\{guerraoui.camelia.kenza.q4,  wang.wenzhi.r7, naito.shoichi.t1\}@dc.tohoku.ac.jp} \\
  \small
 \texttt{jungmin.choi@riken.jp} \quad
 \texttt{kentaro.inui@mbzuai.ac.ae}
 }

\begin{document}
\nolinenumbers
{\makeatletter\acl@finalcopytrue
 \maketitle
}

\begin{abstract}
Prior research in computational argumentation has mainly focused on scoring the quality of arguments, with less attention on explicating logical errors. In this work, we introduce four sets of explainable templates for common informal logical fallacies designed to explicate a fallacy's implicit logic. Using our templates, we conduct an annotation study on top of 400 fallacious arguments taken from LOGIC dataset and achieve a high agreement score (Krippendorf's $\alpha$ of 0.54) and reasonable coverage (0.83). Finally, we conduct an experiment for detecting the structure of fallacies and discover that state-of-the-art language models struggle with detecting fallacy templates (0.47 accuracy). To facilitate research on fallacies, we make our dataset and guidelines publicly available.

\end{abstract}

\section{Introduction}

A \emph{fallacy} is an invalid or weak argument supported by unsound reasoning~\cite{Hinton2020EvaluatingArgument}.
The automatic detection of fallacies has important applications, including providing constructive feedback to learners in writing.
The assessment of argument quality, including fallacy detection, is considered an important topic in the fields of computational argumentation and argumentation mining~\cite{wachsmuth-etal-2017-computational,ijcai2019p879}.

\begin{figure}[t!]
    \centering
    \includegraphics[width=\linewidth]{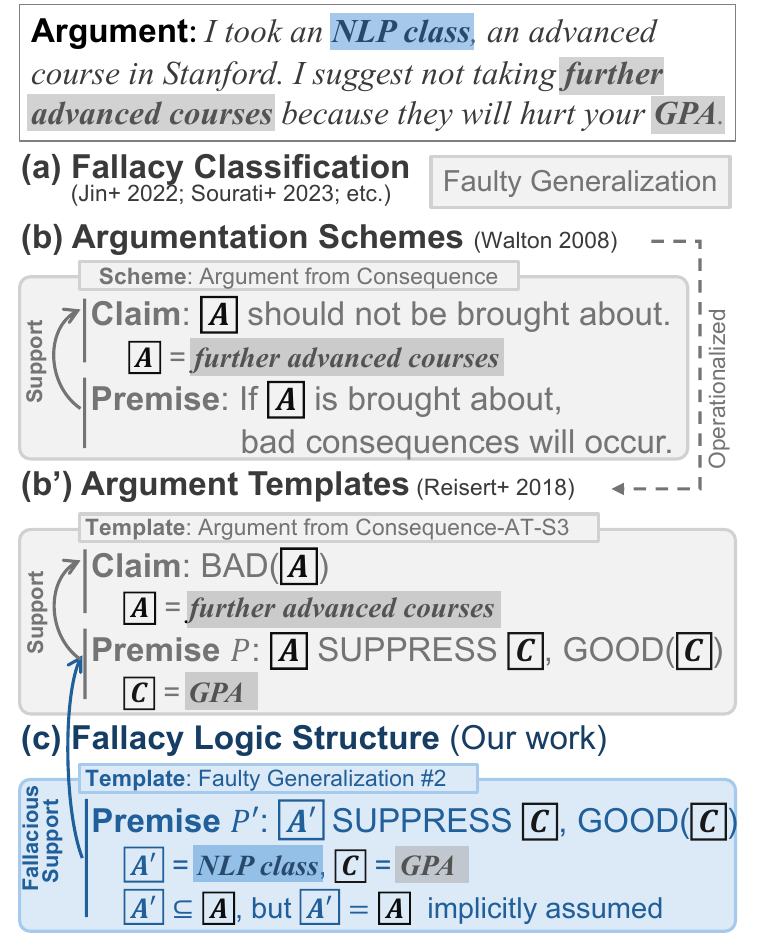}
    \caption{Overview of our proposed fallacy logic structure. We extend (b') existing argumentative representation~\cite{reisert2018feasible} consisting of \hlcolor{\prevwcolor}{Claim} and \hlcolor{\prevwcolor}{Premise $P$} by adding (c) \hlcolor{\proposedwcolor}{Premise $P'$}, which explains what makes the argument fallacious.
    The example annotation shows: (i) the claim ``\hlcolor{\prevwcolor}($A'$={\emph{further advanced courses}) are BAD}'' is supported by ``\hlcolor{\prevwcolor}{$P$: ($A$=\emph{further advanced courses}) SUPPRESS ($C$= \emph{GPA}), a GOOD thing}'', and (ii) $P$ is then further supported by  ``\hlcolor{\proposedwcolor}{$P'$: ($A'$=\emph{NLP class}) SUPPRESS ($C$=\emph{GPA}), a GOOD thing}'', where  $A'$=\hlcolor{\proposedwcolor}{\emph{NLP class}}(their own experience) is implicitly generalized to $A$=\hlcolor{\prevwcolor}{\emph{further advanced courses}}(advanced courses in general), which makes the overall argument fallacious.
    }
     \label{fig:overview}
\end{figure}

Previous work on quality assessment has focused on numerical scoring~\cite{carlile-etal-2018-give,ke-etal-2019-give} and fallacy type-labeling tasks~\cite{jin-etal-2022-logical,Sourati2023Case-BasedFallacies}, without aiming to analyze \emph{fallacy logic structures}, namely the representation of \emph{how} given arguments are weak.
In the field of argumentation theory, a typology of invalid arguments has been long studied and compiled into an inventory~\cite{walton1987informal,bennett2012logically}.
The inventory typically includes semi-formal definitions and some examples for each type of fallacy.
For example, \emph{Faulty Generalization} is a widely recognized fallacy type, characterized by ``Drawing a conclusion based on a small sample size, rather than looking at statistics that are much more in line with the typical or average situation.''~\cite{bennett2012logically}.
The semi-formal definition is as follows: ``(i) Sample $S$ is taken from population $P$. (ii) Sample $S$ is a very small part of population $P$. (iii) Conclusion $C$ is drawn from sample $S$ and applied to population $P$''.
Although such inventory provides insights into how the analysis of fallacy logic structure can be formulated as an NLP task, several important questions remain: (i) How should the annotation scheme for fallacy logic structure identification be designed? (ii) Can humans consistently annotate fallacy logic structures? (iii) To what extent is the automatic identification of fallacy logic structure a challenging task for machines?

To address this issue, we propose \emph{fallacy logic structure identification}, a new task for identifying the underlying logical structure of fallacies.
For this task, we design an annotation scheme and conduct an annotation study to examine its feasibility.
The key idea behind our annotation scheme is to enrich previous work on the argumentative structure with a fallacy structure from an inventory of common fallacy types.

Consider the argument in Fig.~\ref{fig:overview}, where the writer persuades people \emph{not} to take advanced courses at Stanford because they claim it will hurt their GPA. The claim is further supported by the writer's own, single experience based on their NLP class.
This is a faulty generalization caused by the writer \emph{implicitly} assuming that their single experience can be generalized to everyone. Previous work in fallacy identification~\cite{sourati2023robust, jin-etal-2022-logical} would identify this argument as \emph{Faulty Generalization} (Fig.~\ref{fig:overview} (a)), but no additional information such as logical structure or fallacious reasoning is provided. Argumentation Schemes~\cite{walton2008argumentation}, a well-known typology for the representation of arguments, would categorize this argument as \emph{Argument from Consequence} (Fig.~\ref{fig:overview} (b)), and \newcite{reisert2018feasible}'s \textit{Argument Templates}, an operationalized version of Argumentation Schemes, represent this argument with a more fine-grained, logical representation by structured templates (Fig.~\ref{fig:overview} (b')).
To represent the committed fallacy structure, our work further enriches this representation by adding an additional premise that indicates how the given argument is fallacious (Fig.~\ref{fig:overview} (c)).



Our main contributions are as follows:
\begin{itemize}

\item We conduct the first study of formulating logical fallacy structure by creating an inventory of fallacy templates (\S\ref{sec:fls}).


\item We create the first dataset of fallacy logical structures which consists of 400 arguments from LOGIC~\cite{jin-etal-2022-logical} annotated with our templates (\S\ref{sec:dataset}). We publicly release both the dataset and guidelines\footnote{\url{https://github.com/itsanonnymous/fallacytemplate}}. Our dataset achieves high inter-annotator agreement (Krippendorf's $\alpha$ of 0.54) and coverage (0.83\%).

\item We show that the fallacy logic structure identification task poses a significant challenge for state-of-the-art language models. (\S\ref{sec:experiments}).
\end{itemize}

\begin{figure*}[t!]
    \centering
    \includegraphics[width=\linewidth]{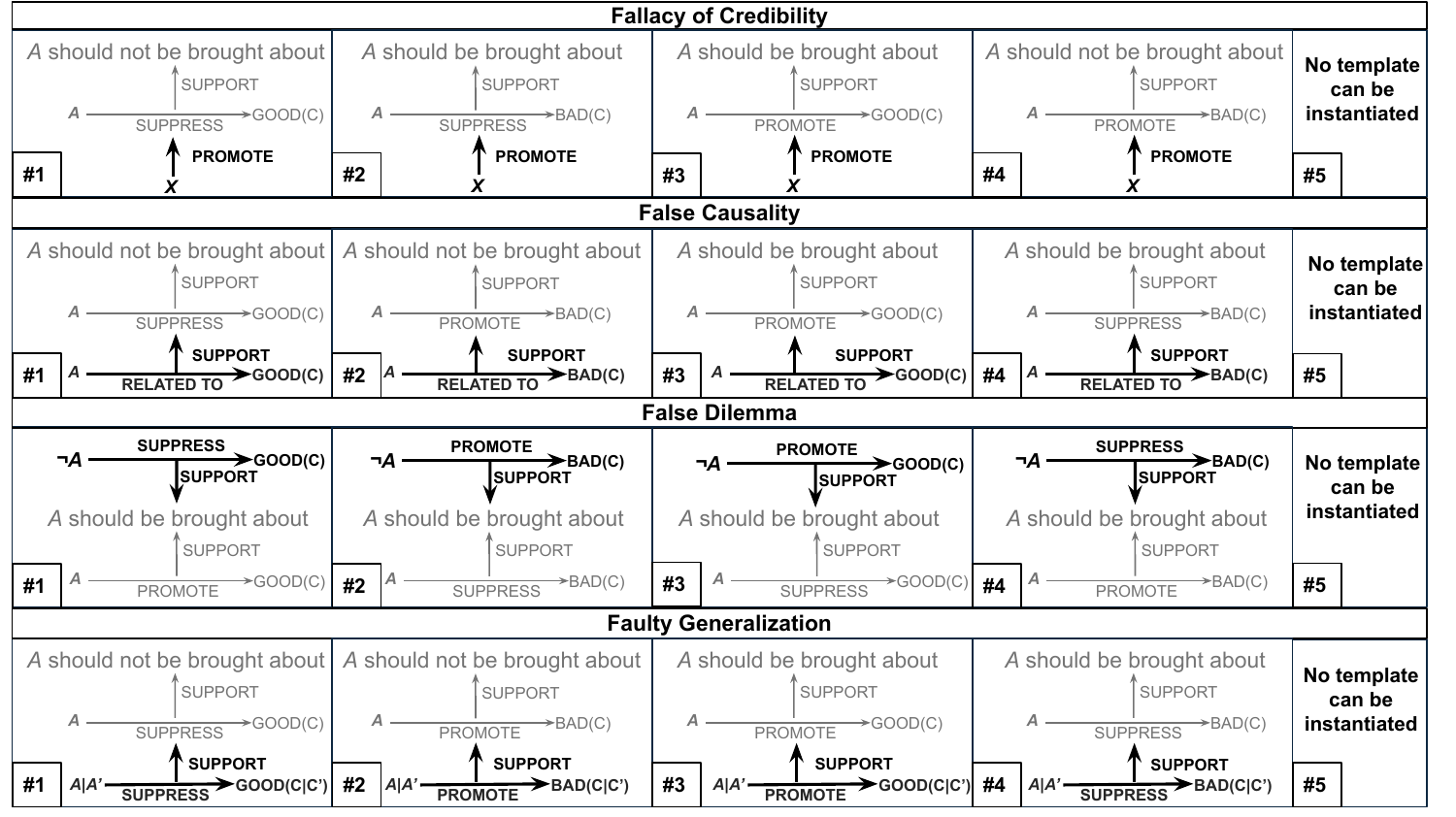}
    \caption{Our templates for annotating fallacious argument logical structure. We extend upon existing work~\cite{walton2008argumentation, reisert2018feasible}, consisting of a conclusion (i.e., \textit{$A$ should (not) be brought about}) and supporting premise, by adding an additional supporting premise in bold which represents the committed fallacy logical structure.}
    \label{fig:templates}
\end{figure*}

\section{Related Work}

\paragraph{\textbf{Fallacies Annotation Study}}


Several studies address creating benchmarks for fallacy identification, including \cite{habernal-etal-2017-argotario} for game facilitation and \cite{ruiz2023detectinginwild} for validating argumentation corpora. Particularly, \newcite{jin-etal-2022-logical} focus on logical fallacies within climate change discourse, emphasizing the challenges posed by complex scientific data. They developed detailed annotation guidelines to aid in consistent identification of fallacies within climate arguments. Similarly, \newcite{goffredo2023argument} analyzed fallacious reasoning in U.S. presidential debates, highlighting common fallacies. They employed advanced computational techniques and the INCEpTION platform for structured annotation, ensuring reliability through cross-verification and Krippendorff's $\alpha$. In addition to the current benchmark establishment, this research proposes benchmark resources aimed at capturing fallacy structure rather than solely identifying fallacies. This research fills the gap, extending previous work by focusing on template annotation to capture the underlying structure of fallacious arguments.


\paragraph{Argumentation Structure}

Argumentation theory examines how arguments, including those about daily exercise, are constructed and evaluated. To begin with, \cite{stab2017parsing} establishes methods for parsing argumentation structure in persuasive essays by identifying and classifying argument components and their relationships. \cite{toulminModel2003uses} provides a framework for analyzing arguments by breaking them down into components like Claim, Grounds, Warrant, and Rebuttal. \cite{walton2013argumentation_common} focuses on specific argumentation schemes, such as Argument from Analogy, which compares similar situations to infer outcomes but risks failure with irrelevant similarities (false analogy). The Argument from Consequence \cite{walton2008argumentation} emphasizes potential outcomes of actions, often involving causality and appeals to consequences. Evaluating it requires considering 1) the connection between action and consequence, 2) the quality of supporting evidence, and 3) whether opposing consequences have been addressed. Building on prior work on argument structure, particularly the Argument from Consequence scheme (a frequently used scheme by Walton), this research addresses a gap by using argument templates, inspired by\cite{reisert2018feasible} to capture the structure of fallacies within this scheme. This choice is motivated by the scheme's frequent use and its potential for revealing fallacious arguments. Building on this potential, and inspired by \cite{reisert2018feasible} on templates, we address a gap by using templates to capture the structure of fallacies within the Argument from the Consequence scheme. Previous work on Argument from Consequence demonstrates high coverage in annotation efforts, further supporting this approach.


\section{Fallacy Logic Structure}
\label{sec:fls}

\subsection{Design Principles}



To develop an annotation scheme for fallacy logic structure, we adhere to three key criteria.

First, we require the annotation to be able to explain the underlying structure of fallacy.
We extend the existing representation of arguments (Fig.~\ref{fig:overview} (b')) by an additional premise attached with an explanation as to why it fallaciously supports the original premise (Fig.~\ref{fig:overview} (c)).

Second, our annotation scheme must cover a majority of fallacy types.
We focus on the fallacies most commonly studied in computational argumentation, such as those in \cite{alhindi2023multitask} and \cite{helwe2023mafalda}, whose statistics on fallacy types guide our template design to match the most frequent occurrences. We develop 20 new templates covering four defective induction fallacy types--Fallacy of Credibility, False Causality, False Dilemma, and Faulty Generalization. An example and more detailed explanation regarding four defective induction fallacy types can be seen in section~\ref{sec:defective induction}.

Third, our annotation scheme must utilise \newcite{reisert2018feasible} template selection and slot-filling approach further simplifying annotation while remaining computationally friendly. As inspired by the Argument from Consequence and employing \newcite{reisert2018feasible}'s work as a base scheme, the template design captures both positive and negative consequences within the scheme. This results in two templates for each consequence type, along with a template addressing instances that cannot be directly covered. This approach aims to provide rich information about fallacy structures while simplifying the annotation process.


\subsection{Representation of Core Arguments}
\label{subsec:template_based_formulation}
The underlying structure of arguments has been represented previously with \newcite{walton2008argumentation}'s Argumentation Schemes, a set of roughly 60 schemes which provide structure between argumentative components such as a conclusion (i.e., claim) and premise. An example of a common scheme, Argument from Negative Consequences, is as follows\footnote{For readability, we represent placeholders in brackets.}:

\begin{itemize}
    \item \textbf{Premise ($P$)}: If [$A$] is brought about, bad consequences will plausibly occur.
    \item \textbf{Conclusion}: Therefore, [$A$] should not be brought about.
\end{itemize}
Here, $A$ is a placeholder (i.e., slot-filler)  represents an \textit{action} and $P$ supports conclusion. For the argument in Fig.~\ref{fig:overview}, we represent Argument from Negative Consequence with [$A$]=``further advanced courses''.

Towards operationalizing \newcite{walton2008argumentation}'s Argumentation Schemes into more fine-grained logical representations, \newcite{reisert2018feasible} developed \textit{argument templates}, an inventory of annotation-friendly templates consisting of ingredients such as placeholders. An example of an argument template built on top of Argument from Negative Consequences scheme is as follows:

\begin{itemize}
    \item \textbf{Premise ($P$):} [$A$] \textit{SUPPRESS} a \textit{GOOD} [$C$].
    \item \textbf{Conclusion}: [$A$] is \textit{BAD}.
\end{itemize}
Both $A$ and $C$ represent \textit{action} and \textit{consequence} placeholders, respectively. \textit{GOOD} and \textit{BAD} represent the sentiment of each placeholder, and \textit{SUPPRESS} represents the relation between $A$ and $C$, where \textit{SUPPRESS} refers to preventing the consequence~\cite{hashimoto2012excitatory}. 
Revisiting the argument in Fig.~\ref{fig:overview}, we can instantiate the argument template with $A$=``further advanced courses'' and $C$=``GPA''. Such argument templates are a simple, efficient way to represent underlying logic.

As shown for Faulty Generalization fallacies in Figure~\ref{fig:templates}, argument templates were handcrafted to allow for both Argument from Positive Consequence ($A$ \textit{should be brought about}) and Argument from Negative Consequence ($A$ \textit{should not be brought about}) with a supporting $P'$ (grey) consisting of positive (e.g., $A$ PROMOTE GOOD($C$)) and negative (e.g., $A$ SUPPRESS GOOD($C$)) consequences, respectively, where \textit{PROMOTE} refers to the triggering of the consequence~\cite{hashimoto2012excitatory}. We build on top of this for adding logical structure for fallacies.



\subsection{Our Fallacy Template Inventory}

For representing fallacy logical structure, we extend \newcite{walton2008argumentation} and \newcite{reisert2018feasible} by introducing a new premise $P'$ which supports premise $P$. Consider the following representation for Faulty Generalization:


\begin{itemize}
    \item \textbf{Premise ($P$):} [$A$] \textit{SUPPRESS} a \textit{GOOD} [$C$].
     \item \textbf{Premise (P'):} [$A'$], a subset of $A$, \textit{SUPPRESS} a \textit{GOOD} [$C$]
    \item \textbf{Conclusion}: [$A$] is \textit{BAD}.
\end{itemize}
Here, on top of the argument template placeholders $A$ and $C$, $P'$ includes a new placeholder $A'$, where $A'$ is an action and $A'\subseteq{A}$. The faulty generalization is committed as a result of the argument considering $A'$ to represent $A$ as a whole.Revisiting the argument in Fig.~\ref{fig:overview}, we can instantiate the above with $A$=``further advanced courses'', $A'$=``NLP class'', and $C$=``GPA''.


\begin{figure}[t]
    \centering
\includegraphics[width=0.85\linewidth]{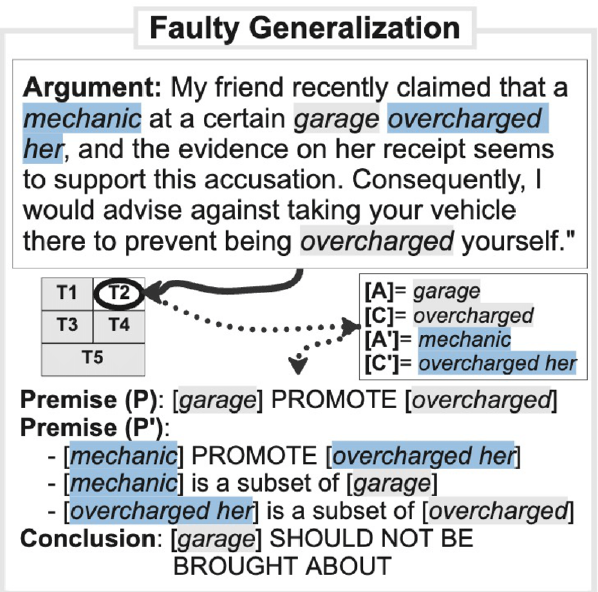}
    \caption{Examples of template and slot-fillers from FtF for Faulty Generalization.}
    \label{fig:examples}
\end{figure}




Fig.~\ref{fig:examples} shows additional examples of template instantiation with placeholders for each target fallacy type, with our new premise $P'$. Using this figure, we exemplify a complex Faulty Generalization argument, where two subsets $A'$ and $C'$ are considered. The main point is symbolized by $A$=``garage'' and $C$=``overcharged'', as the narrative implies that the $A$ is notorious for $C$. Hence, it is implicated that $C$ is \textit{BAD} and that $A$] \textit{PROMOTE} $C$. 
In $P'$,' $A'$=``mechanic'' and $C'$=``overcharged her'' are identified, where $A'\subseteq{A}$ and $C'\subseteq{C}$ and $A'$ PROMOTE $C'$. Therefore, the relation $A'$ PROMOTES $C'$ supports the relation $A$ PROMOTE $C$, so template \#2 is selected.

\section{Flee the Flaw (FtF) Dataset}
\label{sec:dataset}
We discuss the creation of our dataset \textit{Flee the Flaw} (henceforth, \textit{FtF}). First, we use an existing dataset of annotated fallacious arguments for creating our guidelines and building our inventory of fallacy templates. We then conduct a full-fledged annotation on top of 400 arguments.

\subsection{Data Collection}

To build a dataset of fallacious argument template instantiations, we require fallacious arguments which cover our target fallacy types. Therefore, we use LOGIC~\cite{jin-etal-2022-logical}, an English fallacy dataset consisting of 2,449 fallacious arguments spanned across multiple fallacy types, including our four target template types. We sampled 400 arguments (100 per target fallacy type) from LOGIC, equally split between its development (LOGIC-DEV$_{200}$) and training sets (LOGIC-TRAIN$_{200}$), with 200 arguments each. Missing fallacy instances in the development set were supplemented from the training set, ensuring no overlap by segmenting the training set before distribution.

\begin{figure*}[!t]
    \centering
    \includegraphics[width=\textwidth]{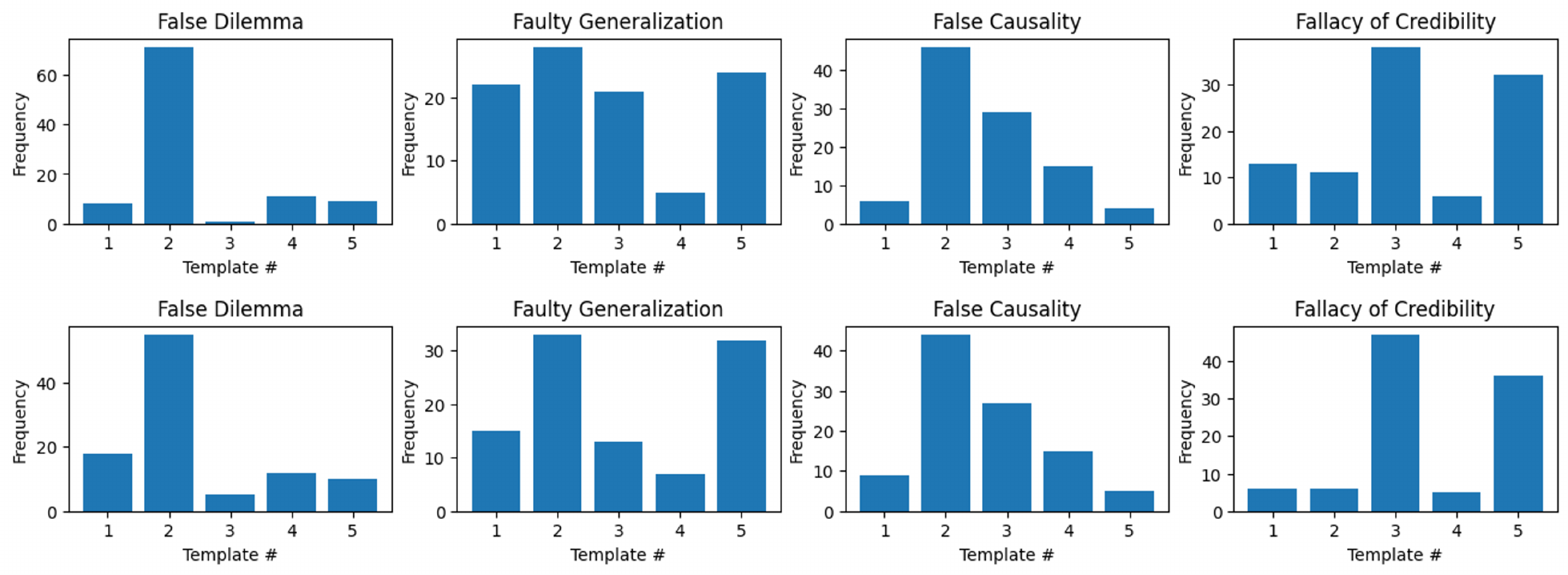}
    \caption{The distribution of fallacy templates in our FtF between one annotator (top row) and the other (bottom row) for all 400 instances in our train and dev set, where each fallacy type consists of 100 instances. The x-axis refers to the selected template, and y-axis refers to the frequency.}
    \label{fig:distribution}
\end{figure*}

 \subsection{Guideline Construction} 

We employed two expert annotators for guideline development and annotation: a native English-speaking postdoctoral researcher specializing in argument mining (who led guideline creation), and a non-native English-speaking graduate student specializing in argumentation. 

To create a set of guidelines and test annotation feasibility, we conduct a multi-round pilot study on top of LOGIC-DEV$_{200}$. Aside from the pilot study itself, annotators did not go through any training phase. Given that the LOGIC dataset has limited fallacious arguments, our pilot study consisted of 200 instances (50 per fallacy type) for creating our final guidelines, where the study began with an initial set of guidelines for all fallacy types. For each of the four fallacy types, annotators focused on the 50 instances per each fallacy. For each type, we split up the instances to annotate (e.g., 10 out of 50) using the latest updated set of guidelines, where results were compared and discussed after each round. Discussion consisted of findings and whether annotators agree with each other’s annotation. If there was a new finding or disagreement, instances were discussed to reach a consensus and guidelines were updated accordingly. The process was repeated until all 200 instances in LOGIC-DEV$_{200}$ were annotated and the final annotation guidelines were created.\footnote{The final guidelines are made publicly available: \url{https://github.com/itsanonnymous/fallacytemplate}}

\paragraph{Reducing Annotation Complexities}
During guideline construction, annotators found that multiple templates could be instantiated for a single argument. In order to reduce annotation complexity, the following conditions were created: i) \textit{preservation of argument's original, explicit intent}, ii) \textit{paraphrase arguments into Argument from Consequences}, and iii) \textit{preference of entities over events.} 

We demonstrate such conditions with the False Dilemma argument: ``\emph{We either have to cut taxes or leave a huge debt for our children.}''. Opposed to selecting the entity $A$=``\textit{taxes}'' which satisfies the third condition, annotators were encouraged to select the event $A$=``\textit{cut taxes}'' as it maintains the explicit intention of the argument, satisfying the first condition. Given that this is a \textit{False Dilemma} fallacious argument which follows an \textit{either-or}, the annotators satisfied the second condition by considering that the argument can be thought of in terms of argument from consequence, where the conclusion ``cut taxes should be brought about'' is good as it suppresses the premise ``leave a huge debt for our children'', a bad thing.

In addition to the above, it was discovered that the fallacy type provided by LOGIC could be categorized into other, non-target fallacy types (e.g., \textit{Slippery Slope} instead of \textit{Faulty Generalization}). In such instances, annotators were instructed to annotate the instance considering its given type and encouraged to apply template \#5 if the template instantiation could not be made.

\subsection{Annotation Procedure}
Given a fallacious argument, its fallacy type, and our templates, the procedure for fallacious template instantiation is as follows. First, annotators select the appropriate template from the given set of 5 templates. Next, annotators write in the necessary slot-fillers taken from the input argument. Afterwards, annotators provide their confidence level for instances in which they are not 100\% confident. Finally, annotators provide any necessary comments to accompany the annotation. The resulting annotation of our fallacious templates on top of LOGIC-DEV$_{200}$ and LOGIC-TRAIN$_{200}$ resulted in FtF-DEV and FtF-TRAIN, respectively.

\begin{table}[t]
    \small
    \begin{tabular}{lr>{\raggedleft\arraybackslash}p{1.8cm}%
    }
    \toprule
     \textbf{Fallacy Type} & \textbf{GWET AC1} & \textbf{Krippendorff's $\alpha$} \\
    \midrule
        False Dilemma  &0.63&0.44\\
        Faulty Generalization &0.40 &0.36\\
        False Causality  & 0.71 &0.65\\
        Fallacy of Credibility  & 0.58&0.49\\        
    \midrule
    Average &  0.57&0.54\\
    \bottomrule
    \end{tabular}
         \caption{Template selection Inter-Annotator Agreement.}
    \label{tab:final_agreement}
\end{table}


\subsection{Statistics and Analysis}

\paragraph{Inter-Annotator Agreement (IAA)}
Table~\ref{tab:final_agreement} shows our IAA scores for template selection. Our GWET AC1~\cite{gwet2008computing} scores range from 0.40 to 0.71, indicating moderate to the substantial agreement. We also calculate Krippendorff’s alpha~\cite{hayes2007answering} and achieve a score of 0.54, indicating a high agreement.

Given that \textit{Faulty Generalization} had the lowest agreement, we conduct an additional analysis on all disagreements for \textit{Faulty Generalization} arguments. We discover that 60\% of disagreements were caused when one annotator labeled '\#5' and the other instantiated a template, where reasons annotators labeled '\#5' were due to complicated instances and implicitness of the argument. Lastly, some instances in LOGIC were found to be other types of fallacies, namely \textit{Slippery Slope}.

\paragraph{Distribution of Templates}
Fig.~\ref{fig:distribution} shows the distribution of the fallacy templates for both annotators. We immediately observe that certain templates were rarely selected by annotators for LOGIC, such as template \#3 for False Dilemma. Regardless of this skewed distribution, as reported, we still achieved a high IAA and coverage for template selection. 


\paragraph{Coverage}
Table~\ref{tab:coverage} provides a comparison of annotation coverage forannotators, namely the percentage of instances where a non-template~\#5 is annotated. Overall, our templates achieve a high coverage for both annotators, with scores of 80\% and 83\%. We observe that fallacy types such as \textit{False Dilemma} and \textit{False Causality} achieve high coverage due to their straightforward reasoning.



\begin{table}[t]
    \centering
    \small
    \begin{tabular}{lrr}
    \toprule
     \textbf{Fallacy Type} & \textbf{Annotator 1}  & \textbf{Annotator 2}\\
    \midrule
        False Dilemma  & 0.90 & 0.91\\
        Faulty Generalization & 0.68 & 0.76\\
        False Causality  & 0.95 & 0.96 \\
        Fallacy of Credibility  & 0.64 & 0.83\\
    \midrule
    Average & 0.80 & 0.83 \\
    \bottomrule
    \end{tabular}
         \caption{Coverage of fallacy templates for both annotators.}
    \label{tab:coverage}
\end{table}

\section{Experiments}
\label{sec:experiments}
To what extent is the automatic identification of fallacy logic structure challenging for machines? We evaluate current state-of-the-art LLMs for FtF.

\subsection{Methodology}
The fallacy logic identification task comprises two sub-tasks: (i)~\textit{template selection} and (ii)~\textit{slot-filling}.  As shown in Table~\ref{tab:generalized_prompt}, the prompt includes this fallacy-type information, allowing LLM to focus on two key actions. In template selection, the model chooses the template that best reflects the fallacious structure.  For slot-filling, the model fills in the slots of the selected template.


It is commonly known that dataset creation in argumentation requires significant resources (human, time, financial), making it difficult to acquire highly reliable large-scale annotations.
Therefore, we employ LLMs with in-context learning and fine-tuning to model both sub-tasks jointly.
We experiment with three distinct prompts: (i) NL$_1$, a pure natural language prompt, (ii) NL$_2$, simplified version of NL$_1$, and (iii) PL, a semi-structured prompt with propositional logic and mathematical notation.
Table~\ref{tab:generalized_prompt} summarizes a general form of these prompts; see Appendix~\ref{app:prompt} for an example of the 1-shot prompt for False Dilemma.\footnote{Detailed prompts used in our experiments are publicly available at \url{https://github.com/itsanonnymous/fallacytemplate/tree/main/ftf_prompts}}

\begin{table}[t]
    \small
    \begin{tabular}{p{0.94\columnwidth}}
    \toprule
    \# \textbf{Task}\\
    Identify the underlying structure of an argument of \{fallacy\_type\}.\\
    Given a list of fallacy templates, your task is to choose a template that best describes the underlying fallacy structure...\\
    \# \textbf{List of Templates}\\
    Template No.1:\escape{n} \{template\_1\}\\
    ...\\
    Template No.5:\escape{n}There is either no consequence in the argument.\\
    \# \textbf{Output Format}\\
    Template No.=[No.]\escape{n}\{slot\_fillers\}\\
    \# \textbf{Example}\\
    \{examples\}\escape{n}\# \\
    \# \textbf{Query}\\
    \{\}\\\hline
    \end{tabular}
    \caption{Generalized prompt used for our 0, 1, and 5-shot LLM experiments. \{fallacy\_type\} is either Fallacy of Credibility, False Causality, Faulty Generalization, or False Dilemma. Depending on the fallacy type, the appropriate templates and slot-filler choices are provided to the prompt, and for 1 and 5-shot settings, \{examples\} are provided. For spacing purposes, we replace newlines with \escape{n} in this prompt and omit templates 2-4.}
    \label{tab:generalized_prompt}
\end{table}


\subsection{Setup}




\paragraph{Models}
We employed four state-of-the-art LLMs: GPT-3.5-turbo~\cite{abdullah2022chatgpt}, GPT-4o~\cite{achiam2023gpt}, Llama-3-8B\cite{meta2024llama}, and Mistral-7B\cite{jiang2023mistral}. We use a temperature of 0, max tokens of 0.6, top\_p of 1.0, and both frequency and presence penalties of 0. Experiments were conducted using zero-shot, one-shot, and five-shot prompt settings for GPT-3.5-turbo and GPT-4o.  Few-shot examples were sampled from FtF-TRAIN, with the number of shots reflecting the number of examples provided in the prompt.

For the fine-tuned model, we split Ftf-TRAIN into 150 instances as training data and 50 instances as validation data. We set the learning rate into 2e-4 and optimizer adamw8bit. All models used Ftf-DEV for testing and evaluating the results. 

\paragraph{Evaluation Metrics}
We use accuracy for the template section. For the slot-filling, we will target only instances where the template is correctly identified by the model. Formally, we define \textit{exact-match slot-filling accuracy} as follows: $\frac{|X \cap Y|}{|X|}$, where $X$ is a set of test instances where the predicted template is correct, and $Y$ is a set of test instances where \emph{all} predicted slot-fillers must exactly match the gold-standard slot-fillers.\footnote{We lowercased all tokens for word matching.}
In addition, we use \textit{partial-match slot-filling accuracy}, where $Y$ is a set of test instances where \emph{all} predicted slot-fillers are required to have over 50\% word overlap with the gold standard.

For evaluating overall performance, we define a \emph{joint accuracy} to be a multiplication of template selection accuracy and slot-filling accuracy.

\subsection{Results and Analysis}

Tables~\ref{tab:bestprompt_fine_tune} demonstrate low accuracies across all models. We choose NL$_2$ prompt for comparing the result with fine-tuned models based on the highest accuracy in the template selection between GPT4 and GPT3.5 (appendix~\ref{prompt_type_ts_em}). Regarding template selection, the Mistral-7B model generally outperforms every model. Conversely, in slot-filling, the results show that the GPT4 models with 5-shot prompting outperform every model. Overall, the low joint accuracy highlights a significant limitation of state-of-the-art language models in identifying the logical fallacy structure that best captures the underlying fallacious structure within FtF.
Improving LLMs' ability to handle slot-filling tasks remains a significant challenge.

\subsection{Error Analysis}\label{sec:error}


We conducted an error analysis on 40 instances, aiming to improve the template and prompt. We focused on the Mistral-7b's generated results due to their highest joint accuracy. We discovered the following errors: (i)The model predicted template 5 despite the argument being able to be instantiated (32.5\%), (ii)The model predicted a different template due to different slot-fillers (32.5\%), (iii)The model predicted a different template despite having similar slot-fillers as the gold label (17.5\%), and (iv)The model instantiated the template despite no argument from consequence (17.5\%). 

We found that template 5 was sometimes predicted due to noise in the input argument. Among all the instances that fell into category (i), four instances predicted template 5 because of this noise.\footnote{See Example 1 in Appendix Table~\ref{tab:error} for an example of input question noise leading to template 5}

Although the prompts were built off our guidelines, we found that the model occasionally selected different templates due to many possible terms for slot-filling. Example 2 in table~\ref{tab:error}, shows that the model selects a different template due to the difference in slot-filler $A$. Upon further analysis, the model's predicted answer was also correct, as ``ban hairspray'' suppress ``the world will end'' possesses the same semantic meaning as ``hairspray'' promotes ``the world will end'', but the selected template and slot-filler were both incorrect.


It still remains a question why such an error that falls into categories (iii) and (iv) occurred. However, we believe that the performance drop in the model was attributed to three main factors: noise in the dataset, the presence of multiple templates that could be selected, and the existence of various possible terms that could fill the slot-filler. Further details can be seen in section~\ref{sec:error_further}.



\begin{table}[t!]\centering
\footnotesize
\begin{tabular}{lccc}\toprule
\textbf{Model} &\textbf{Acc. (TS)} &\textbf{Acc. (SF)} &\textbf{Acc. (Joint)} \\\midrule
GPT4NL$_2$-0 &0.36 &0.06 &0.02 \\
GPT4NL$_2$-1 &0.42 &0.10 &0.04 \\
GPT4NL$_2$-5 &0.38 &\textbf{0.24} &0.09 \\
\midrule
GPT3.5NL$_2$-0 &0.21 &0.06 &0.01 \\
GPT3.5NL$_2$-1 &0.30 &0.14 &0.04 \\
GPT3.5NL$_2$-5 &0.35 &0.17 &0.06 \\
\midrule
Llama3-7b &0.34 &0.16 &0.05 \\
Mistral-7b &\textbf{0.47} &0.23 &\textbf{0.11} \\
\bottomrule
\end{tabular}
\caption{Model accuracy for template selection (TS) and exact-match accuracy for slot-filling (SF). }\label{tab:bestprompt_fine_tune}
\end{table}

\section{Conclusion and Future Work}
\label{sec:conclusion}


In this work, we conduct the first study to address logical fallacy structure by creating an inventory of fallacy templates. In total, we created 20 novel templates spanned across 4 fallacy types (Fallacy of Credibility, False Causality, False Dilemma, and Faulty Generalization). We created and released Flee the Flaw, a new dataset consisting of 400 arguments from LOGIC~\cite{jin-etal-2022-logical} annotated with fallacy logic structure and publicly released both the corpus and guidelines. Our dataset achieved a high inter-annotator agreement (Krippendorf's $\alpha$ of 0.54) and coverage (0.83\%). We experiment on top of our new dataset by conducting In-Context Learning and fine-tuning for fallacy logic structure identification and discover that it is still a significant challenge for state-of-the-art language models.

Our next step involves studying the underlying patterns and reasoning errors in arguments by analyzing the logical structure of fallacies. Simultaneously, we plan to conduct large-scale annotation on top of lengthier, more natural arguments. Finally, we plan to explore non-consequential topics, allowing for more Argumentation Schemes to be considered.

\section*{Limitations}


In this research, we mainly focus on the proposed explainable fallacy template for only 4 fallacy types which are all mainly informal fallacies. We do not address the fallacy of logic which is the extension from the informal fallacy to formal fallacy. To keep annotation simple, our fallacy templates do not cover every possible combination of ingredients (e.g. relations such as \textit{NOT PROMOTE}, \textit{NOT SUPPRESS}) which limits the amount of total instantiations we can acquire. Regardless, we still achieved a coverage score of roughly 80\%. Furthermore, we extend on argument templates~\cite{reisert2018feasible} which were inspired by \newcite{walton2008informal}'s Argument from Consequence scheme which is a common scheme for every day arguments, but may limit the full range of fallacy instantiations that we can produce.

We limit ourselves to four types of fallacies which only represents a small subset of all known fallacies. Primarily, we target common informal logical fallacies as a start for fallacious template structure instantiation.
Given the structure of \textit{False Dilemma} fallacy, which follows an \textit{either-or} structure, we obtain an unbalanced partition for our False Dilemma templates. As shown in Fig.~\ref{fig:distribution}, both annotators mainly annotated with template 2.

\section*{Ethical Considerations}

Each author of this paper ensured that all ethical considerations were upheld. All results are reported as accurately as possible. Given that we conducted an annotation, we adhere to constructing a high quality dataset as exemplified by our annotator agreement results.




\bibliography{anthology,custom}
\bibliographystyle{acl_natbib}

\clearpage

\appendix
\section{Appendix}
\label{sec:appendix}





\subsection{Template Examples}

\begin{figure}[h]
    \centering
\includegraphics[width=\linewidth]{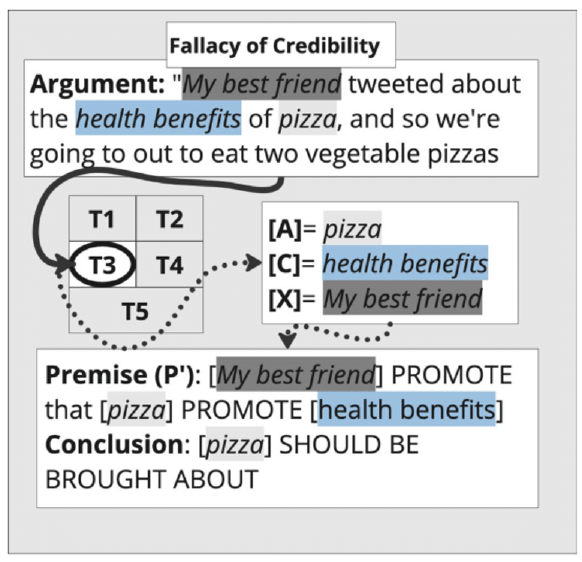}
    \caption{Examples of template and slot-fillers from FtF for Fallacy of Credibility.}
    \label{fig:examples_cred}
\end{figure}

Shown in Fig.~\ref{fig:examples_cred} is the example of Fallacy of Credibility. For the Fallacy of Credibility argument, the fallacy is committed as the $X$=``best friend'' is promoting that $A$=``pizza'' has $C$=``health benefits'', resulting in $P'$: $X$=``best friend'' promote that $A$=``pizza'' promote $C$=``health benefits'', thus Conclusion is $A$=``pizza'' should be brought about. However, the friend is not an expert in the field of nutrition.

\begin{figure}[h]
    \centering
\includegraphics[width=\linewidth]{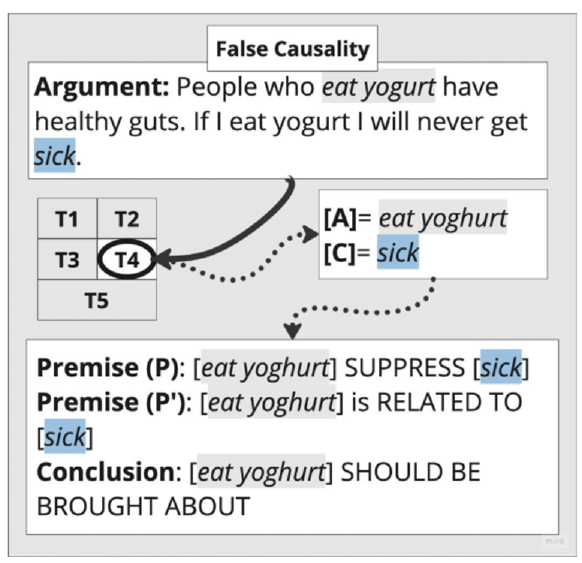}
    \caption{Examples of template and slot-fillers from FtF for False Causality.}
    \label{fig:examples_c}
\end{figure}

For the False Causality argument shown in Fig.~\ref{fig:examples_c}, the argument is stating that $A$=``eat yoghurt'' has a correlation with people with healthy guts, and thus the $P$: $A$=``eat yoghurt'' suppressing $C$=``sick''. The False Causality is linked, as it's implying that $A$=``eating yoghurt'' will definitely suppress $C$=``sick''. In conclusion, $A$=``eating yoghurt'' should be brought about.

\begin{figure}[h]
    \centering
\includegraphics[width=\linewidth]{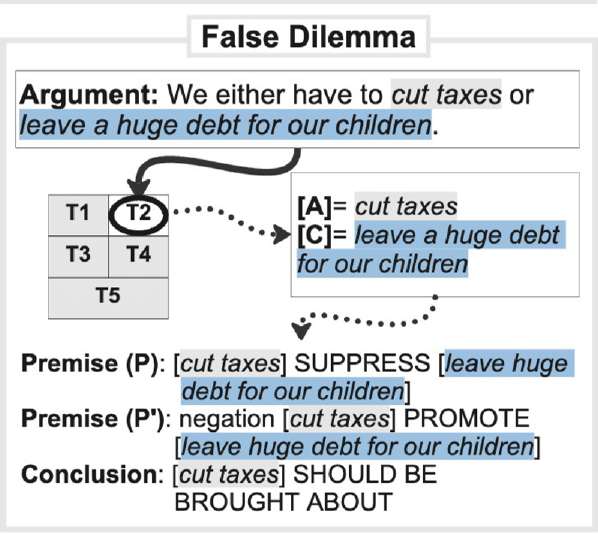}
    \caption{Examples of template and slot-fillers from FtF for False Dilemma.}
    \label{fig:examples_d}
\end{figure}

The example of argument shown in Fig.~\ref{fig:examples_d} is considered as False Dilemma fallacy. The argument limited the option to $A$=``cut taxes'' and negation of $A$=``cut taxes'' for determine the consequence of $C$=``leave a huge debt for our children''. It conclude $A$=``cut taxes'' should be brought about without considering any possible action except $P$: $A$=``cut taxes'' suppress $C$=``leave a huge debt for our children'', and $P'$: negation of $A$=``cut taxes'' promote $C$=``leave a huge debt for our children''. 

\subsection{Fallacy Types}\label{sec:defective induction}
\textit{False Dilemma} occurs due to the restriction of the choices and ignoring additional potential options. \textit{Faulty Generalization} occurs when a belief is applied to a large population without a sufficient and unbiased sample. \textit{False Causality}  assumes a cause-and-effect relationship between two events. Finally, \textit{Fallacy of Credibility} involves an appeal to ethics, authority, or credibility that is not directly relevant to the argument.
Table~\ref{tab:definition_example} provides a definition, example, and further explanation of the example for False Dilemma, Faulty Generalization, False Causality, and Fallacy of Credibility.

\begin{table*}[t!]\centering
\footnotesize
\begin{tabular}{p{20mm}p{45mm}p{40mm}p{45mm}}\toprule
\textbf{Fallacy Type} &\textbf{Definition} &\textbf{Example} &\textbf{Explanation} \\\midrule
\begin{tabular}[t]{p{20mm}}
     False Dilemma
\end{tabular}&
\begin{tabular}[t]{p{40mm}}
     This fallacy is when incorrect limitations are made on the possible options in a scenario when there could be other options.
\end{tabular}&
\begin{tabular}[t]{p{35mm}}
     We either have to cut taxes or leave a huge debt for our children
\end{tabular}&
\begin{tabular}[t]{p{40mm}}
     This argument only limits the options into ``cut taxes'' or ``not cut taxes'' for dealing with a ``debt'' without considering other potential options. 
\end{tabular}
\\
\hline
\begin{tabular}[t]{p{20mm}}
     Faulty \\Generalization 
\end{tabular}&
\begin{tabular}[t]{p{40mm}}
     This fallacy occurs when an argument applies a belief to a large population without having a large enough sample to do so. 
\end{tabular}&
\begin{tabular}[t]{p{35mm}}
     I took an NLP class, an advanced course in Stanford. I suggest not taking further advanced courses because they will hurt your GPA. 
\end{tabular}&
\begin{tabular}[t]{p{40mm}}
     This argument generalizes ``further advanced courses'' should not be taken due to hurting the person's "GPA" only because the took one of the advanced courses ``NLP class.''
\end{tabular}
\\
\hline
\begin{tabular}[t]{p{20mm}}
     False Causality 
\end{tabular}&
\begin{tabular}[t]{p{40mm}}
     This fallacy occurs when an argument assumes that since two events are correlated,they must also have a cause and effect relationship. 
\end{tabular}&
\begin{tabular}[t]{p{35mm}}
     People who eat yoghurt have healthy guts. If I eat yoghurt I will never get sick  
\end{tabular}&
\begin{tabular}[t]{p{40mm}}
     This argument has a belief that ``eat yoghurt'' has a strong relation with ``never get sick'' because of having ``healthy guts''. Thus, it believes that by eating yoghurt will ``never get sick.''
\end{tabular}
\\
\hline
\begin{tabular}[t]{p{20mm}}
     Fallacy of \\Credibility 
\end{tabular}&
\begin{tabular}[t]{p{40mm}}
     This fallacy is when an appeal is made to some form of ethics, authority, or credibility. 
\end{tabular}&
\begin{tabular}[t]{p{35mm}}
     My Best friend tweeted about the health benefits of pizza, and so we're going to out to eat two vegetable pizzas  
\end{tabular}&
\begin{tabular}[t]{p{40mm}}
     The argument has been promoted by the person's best friend by ``tweet about the health benefits of pizza'', but the person best friend is not an expert in the field of nutrition which makes the argument is not credible.
\end{tabular}
\\
\bottomrule
\end{tabular}
\caption{Definition and example explanation of four defective induction fallacy types }\label{tab:definition_example}
\end{table*}

\subsection{Prompt Type for Template Selection and Exact Match Performance}\label{prompt_type_ts_em}
We report the result of template selection accuracy and the average accuracy of slot-filling for exact match for every three prompt types using GPT-4 in table~\ref{tab:JointAccuracy_GPT4} and GPT3.5 in table~\ref{tab:JointAccuracy_GPT35}. 

Template selection performs better for 1-shot prompting for every prompt type in the GPT4 model. However, not for the slot-filling task, 5-shot prompting outperforms 1-shot prompting for every prompt type despite not having the highest accuracy in the template selection task. Different from GPT3.5 where every task is dominated by 5-shot prompting for every prompt type.

Overall, model performance shows minimal variation based on prompt type, suggesting that prompt variation has no significant impact on performance.

\begin{table}[h!]\centering
\footnotesize
\begin{tabular}{lrrrc}\toprule
\textbf{Pr} &\textbf{$n$} &\textbf{Acc. (TS)} &\textbf{Acc. (SF)} &\textbf{Acc. (Joint)} \\\midrule
NL$_1$ &0 &0.31 &0.10 &0.03 \\
NL$_1$ &1 &\textbf{0.36} &0.12 &0.04 \\
NL$_1$ &5 &0.32 &\textbf{0.22} &\textbf{0.07} \\
\midrule
NL$_2$ &0 &0.36 &0.06 &0.02 \\
NL$_2$ &1 &\textbf{0.42} &0.10 &0.04 \\
NL$_2$ &5 &0.38 &\textbf{0.24} &\textbf{0.09} \\
\midrule
PL &0 &0.32 &0.10 &0.03 \\
PL &1 &\textbf{0.38} &0.10 &0.04 \\
PL &5 &0.31 &\textbf{0.18} &\textbf{0.06} \\
\bottomrule
\end{tabular}
\caption{GPT-4 accuracy for template selection (TS) and exact-match accuracy for slot-filling (SF). $n$ denotes the number of few-shot examples, and  \textbf{Pr} denotes a prompt type. }\label{tab:JointAccuracy_GPT4}
\end{table}

\begin{table}[h!]\centering
\footnotesize
\begin{tabular}{lrrrc}\toprule
\textbf{Pr} &\textbf{$n$} &\textbf{Acc. (TS)} &\textbf{Acc. (SF)} &\textbf{Acc. (Joint)} \\\midrule
NL$_1$ &0 &0.21 &0.12 &0.02 \\
NL$_1$ &1 &0.31 &0.14 &0.04 \\
NL$_1$ &5 &\textbf{0.37} &\textbf{0.19} &\textbf{0.07} \\
\midrule
NL$_2$ &0 &0.21 &0.06 &0.01 \\
NL$_2$ &1 &0.30 &0.14 &0.04 \\
NL$_2$ &5 &\textbf{0.35} &\textbf{0.17} &\textbf{0.06} \\
\midrule
PL &0 &0.21 &0.13 &0.03 \\
PL &1 &0.29 &0.04 &0.01 \\
PL &5 &\textbf{0.37} &\textbf{0.17} &\textbf{0.06} \\
\bottomrule
\end{tabular}
\caption{GPT-3.5 accuracy for template selection (TS) and exact-match accuracy for slot-filling (SF).}\label{tab:JointAccuracy_GPT35}
\end{table}

\begin{table*}[t!]\centering
\footnotesize
\begin{tabular}{p{5mm}p{20mm}p{45mm}p{40mm}p{40mm}}\toprule
&\textbf{Fallacy Type} &\textbf{Example} &\textbf{Correct Answer} &\textbf{Predicted Answer} \\\midrule
\begin{tabular}[t]{p{5mm}}
 1
\end{tabular}&
\begin{tabular}[t]{p{20mm}}
 False Dilemma
\end{tabular}&
\begin{tabular}[t]{p{40mm}}
 ``America: Love it or leave it. This is an example of which kind of logical fallacy?''
\end{tabular}&
\begin{tabular}[t]{p{35mm}}
 Template No.=2\\
 $[A]$=Love it\\
 $[C]$=leave it
\end{tabular}&
\begin{tabular}[t]{p{35mm}}
 Template No.=5\\
 $[A]$=\\
 $[C]$=
\end{tabular}
\\
\hline
\begin{tabular}[t]{p{5mm}}
 2
\end{tabular}&
\begin{tabular}[t]{p{20mm}}
 False Dilemma 
\end{tabular}&
\begin{tabular}[t]{p{40mm}}
 We either ban hairspray or the world will end.
\end{tabular}&
\begin{tabular}[t]{p{35mm}}
 Template No.=4\\
 $[A]$=hairspray\\
 $[C]$=the world will end
\end{tabular}&
\begin{tabular}[t]{p{35mm}}
 Template No.=2\\
 $[A]$=ban hairspray\\
 $[C]$=the world will end\
\end{tabular}
\\
\hline
\begin{tabular}[t]{p{5mm}}
 3
\end{tabular}&
\begin{tabular}[t]{p{20mm}}
 False Causality 
\end{tabular}&
\begin{tabular}[t]{p{40mm}}
 Iâ€™ve never had the flu because I take my vitamins everyday.
\end{tabular}&
\begin{tabular}[t]{p{35mm}}
 Template No.=4\\
 $[A]$=vitamins\\
 $[C]$=flu
\end{tabular}&
\begin{tabular}[t]{p{35mm}}
 Template No.=3\\
 $[A]$=vitamins\\
 $[C]$=flu
\end{tabular}
\\
\hline
\begin{tabular}[t]{p{5mm}}
 4
\end{tabular}&
\begin{tabular}[t]{p{20mm}}
 Faulty \\Generalization 
\end{tabular}&
\begin{tabular}[t]{p{40mm}}
 This new test seemed so promising, but the 3 studies that supported its validity turned out to have critical methodological flaws, so the test is probably not valid.
\end{tabular}&
\begin{tabular}[t]{p{35mm}}
 Template No.=2\\
 $[A]$=test\\
 $[C]$=critical methodolical flaws\\
 $[A']$=3 studies that supported its validity turned out to have critical methodological flaws\\
 $[C']$=
\end{tabular}&
\begin{tabular}[t]{p{35mm}}
 Template No.=5\\
 $[A]$=\\
 $[C]$=\\
 $[A']$=\\
 $[C']$=
\end{tabular}
\\
\hline
\begin{tabular}[t]{p{5mm}}
 5
\end{tabular}&
\begin{tabular}[t]{p{20mm}}
 Fallacy of \\Credibility 
\end{tabular}&
\begin{tabular}[t]{p{40mm}}
 Albert Einstein was extremely impressed with this theory.
\end{tabular}&
\begin{tabular}[t]{p{35mm}}
 Template No.=5\\
 $[A]$=\\
 $[C]$=\\
 $[X]$=
\end{tabular}&
\begin{tabular}[t]{p{35mm}}
 Template No.=2\\
 $[A]$=this theory\\
 $[C]$=Albert Einstein\\
 $[X]$=extremely impressed
\end{tabular}
\\
\bottomrule
\end{tabular}
\caption{False prediction generated by Mistral-7B model}\label{tab:error}
\end{table*}

\subsection{Prompt Type for Template Selection and Partial Match Performance}\label{prompt_type_pm}
We report the average accuracy of slot-filling for partial match. The results are shown in table~\ref{tab:pm_GPT4} for GPT-4 and table~\ref{tab:pm_GPT3} GPT3.5. Despite NL$_2$ zero-shot prompt on GPT4 model performance of only 0.06 accuracy for an exact match slot-filling task in table~\ref{tab:JointAccuracy_GPT4}, it performs the best with 0.49 accuracy in the partial match slot-filling task.

\begin{table}[h!]\centering
\footnotesize
\begin{tabular}{lrrrc}\toprule
\textbf{Pr} &\textbf{$n$} &\textbf{Acc. (TS)} &\textbf{Acc. (SF)} &\textbf{Acc. (Joint)} \\\midrule
NL$_1$ &0 &0.31 &0.24 &0.07 \\
NL$_1$ &1 &\textbf{0.36} &0.43 &\textbf{0.16} \\
NL$_1$ &5 &0.32 &0.32 &0.10 \\
\midrule
NL$_2$ &0 &0.36 &\textbf{0.49} &\textbf{0.17} \\
NL$_2$ &1 &\textbf{0.42} &0.35 &0.15 \\
NL$_2$ &5 &0.38 &0.42 &0.16 \\
\midrule
PL &0 &0.32 &0.32 &\textbf{0.10} \\
PL &1 &\textbf{0.38} &0.21 &0.08 \\
PL &5 &0.31 &\textbf{0.33} &\textbf{0.10} \\
\bottomrule
\end{tabular}
\caption{GPT-4 accuracy for template selection (TS) and partial-match accuracy for slot-filling}\label{tab:pm_GPT4}
\end{table}

\begin{table}[h!]\centering
\footnotesize
\begin{tabular}{lrrrc}\toprule
\textbf{Pr} &\textbf{$n$} &\textbf{Acc. (TS)} &\textbf{Acc. (SF)} &\textbf{Acc. (Joint)} \\\midrule
NL$_1$ &0 &0.21 &0.12 &0.02 \\
NL$_1$ &1 &0.31 &0.33 &0.10 \\
NL$_1$ &5 &\textbf{0.37} &\textbf{0.37} &\textbf{0.14} \\
\midrule
NL$_2$ &0 &0.21 &0.19 &0.04 \\
NL$_2$ &1 &0.29 &\textbf{0.36} &0.11 \\
NL$_2$ &5 &\textbf{0.37} &\textbf{0.36} &\textbf{0.12} \\
\midrule
PL &0 &0.21 &0.20 &0.04 \\
PL &1 &0.30 &\textbf{0.43} &0.12 \\
PL &5 &\textbf{0.35} &0.38 &\textbf{0.14} \\
\bottomrule
\end{tabular}
\caption{GPT-3.5 accuracy for template selection (TS) and partial-match accuracy for slot-filling}\label{tab:pm_GPT3}
\end{table}

\subsection{False Template Prediction}\label{sec:error_further}
Table~\ref{tab:error} provides false prediction results using the Mistral-7B model. As previously mentioned in the section~\ref{sec:error}, we categorize into 4 types of error. We found that one of the reasons is the noise that causes an error in category (i), but such an error also occurred even for non-noisy input like in example 4. 

In example 3, the model correctly predicts the slot-filler but chooses different templates. The template conclusion ``Vitamins should be brought about'' is correct, but the model incorrectly assigns a good sentiment to ``flu'' and creates the premise ``Vitamins promote flu'' which does not align with the argument's intention. It remains unclear whether the model struggles to define sentiment.

In example 5, we believe the argument cannot be instantiated as it is not an argument from consequences. However, the model instantiates this argument into template 2 with the premise ``This theory suppresses Albert Einstein'' and got promoted by ``extremely impressed'' which completely different meaning from the input. This raises the question of whether the model truly understands arguments from consequence and the template structure.

\subsection{Prompt for LLM Experiments}
\label{app:prompt}

Table~\ref{tab:5_shot_false_dilemma_NL1}, Table~\ref{tab:5_shot_false_dilemma_PL}, Table~\ref{tab:5_shot_false_dilemma_NL2} provides an example of the 5-shot prompt for False Dilemma used during our LLM experiments. Instances used for non-zero-shot settings are randomly selected from FtF-TRAIN$_{200}$.

\subsection{Towards Dataset Expansion}

Towards extending FtF with other datasets, using our templates and guidelines, we conduct a preliminary annotation on top of three existing, pre-labeled fallacy datasets: LOGICCLIMATE~\cite{jin-etal-2022-logical}, and Argotario~\cite{habernal-etal-2017-argotario}, and a Covid dataset~\cite{bonial2022search}. We randomly sample 40 arguments, where 20 are labeled as ``Hasty Generalization'' and 20  are labeled as ``Irrelevant Authority''.
We obtain a total coverage of 0.50, with Argotario achieving the highest coverage (0.60) and the Covid dataset achieving the lowest (0.30). Reasons our templates could not be instantiated included instances that \textit{require evidence to function as the target fallacy type} (e.g., ``Covid vaccines contain aborted babies.'' as ``Hasty Generalization''), \textit{not the target type fallacy} (e.g., ``No, because if you start with same sex marriage, what is next? Marriage with animals?'', a Slippery Slope), of no credible source information provided (e.g., ``“The COVID-19 pandemic is not a real medical pandemic”; “The COVID-19 vaccine is not proven safe or effective”'').


\begin{table*}[t]
    \small
    \begin{tabular}{p{170mm}}
    \toprule

    \textbf{\# Task}\\
    Identify the underlying structure of an argument of False Dilemma.\\
    Given a list of fallacy templates, your task is to choose a template that best describes the underlying fallacy structure, choosing the template's placeholders, $[A]$ and $[C]$, directly from the input text. Additionally, the text must be a consecutive sequence of one or more terms without any conjugation. \\
    Please follow the output format.\\
    
    \# \textbf{Definition}s\\
    Entity: a noun phrase in the input.\\
    Event: a verb phrase in the input.\\
    Placeholder: A fill-in-the-blank choice within a template. Each placeholder may either be an entity or an event.\\
    Please note! Placeholders can ONLY be either an entity (i.e., noun phrase) or an event (i.e., verb phrase) and may not be any other type of phrase (e.g., prepositional phrase).\\
    
    \# \textbf{List of Templates}\\
    Template No.1:\\
    Premise 1: An entity/action [A] promotes a good entity/action [C].\\
    Premise 2: The absence of an entity/action [A] suppresses a good entity/action [C].\\
    Conclusion: Therefore, both Premise 1 and Premise 2 support that [A] should be brought about.\\

    Template No.2:\\
    Premise 1: An entity/action [A] suppresses a bad entity/action [C]\\
    Premise 2: The absence of an entity/action [A] promotes a bad entity/action [C].\\
    Conclusion: Therefore, both Premise 1 and Premise 2 support that [A] should be brought about.\\

    Template No.3:\\
    Premise 1: An entity/action [A] suppresses a good entity/action [C]\\
    Premise 2: The absence of an entity/action [A] promotes a good entity/action [C].\\
    Conclusion: Therefore, both Premise 1 and Premise 2 support that [A] should not be brought about.\\

    Template No.4:\\
    Premise 1: An entity/action [A] promotes a bad entity/action [C].\\
    Premise 2: The absence of an entity/action [A] suppresses a bad entity/action [C].\\
    Conclusion: Therefore, both Premise 1 and Premise 2 support that [A] should not be brought about.\\

    Template No.5:\\
    There is either no consequence in the argument, or the argument cannot be instantiated with one of the templates above.\\

    \# \textbf{Output Format}\\
    Template No.=[No.]\\
    $[A]$=\\
    $[C]$=\\

    \# \textbf{Important Criteria}: Prioritizing entities over events for placeholder.\\
    For choosing placeholder, please prioritize entities over events in the case that an entity itself captures the underlying intent of the argument opposed to the event. However, if the event makes more sense, please choose an event for the placeholder.\\

    \# \textbf{Correct Example}\\
    Input: To get better schools, we have to raise taxes. If we don't, we can't have better schools.\\
    Output:\\
    Template No.=1\\
    $[A]$=raise taxes\\
    $[C]$=schools\\
    Explanation:\\
    Here, there are 2 possible options for [C] which are "schools" (i.e., entity) and "can't have better schools" (i.e., event). Since the entity is the top priority and the second option does not work with template 1 because it is a suppressed relation, "schools" is cchosen for [C].\\
    Also, [A] and [C] are taken directly from the input text. For example, "raising taxes" as [A] also sounds correct, but the term "raising" is not mentioned in the input text. That is why "raise taxes" is chosen for [A]. Because the argument believes that "raise taxes" promote "schools" while not "raise taxes" suppress "school". he conclusion is implicit that “Premise 1 supports that raise taxes should be brought about.” Thus, Template No.=1 is selected.\\
    
    \# \textbf{Wrong Example}\\
    Input: To get better schools, we have to raise taxes. If we don't, we can't have better schools.\\
    Output:\\
    Template No.=1\\
    $[A]$=raising taxes\\
    $[C]$=can't have better schools\\
    Explanation:\\
    Here, there are 2 possible options for [C] which are "schools" (i.e., entity) and "can't have better schools" (i.e., event). However, "can't have better schools" as [C] is incorrect because it is an event instead of the entity of "schools" which already makes sense.\\
    Also, "raising taxes" as [A] is incorrect because the placeholder is not taken directly from the text. Here "raising taxes" is chosen as [A] but the word "raising" does not appear in the input text. Therefore the correct choice for [A] is "raise taxes". \\

    \# Example1\\
    If you can't prove that Ken had an affair with the nanny, then he's been faithful to his wife.\\
    Template No.=1\\
    $[A]$=prove that Ken had an affair with the nanny\\
    $[C]$=he's been faithful to his wife\\

    Again, please only select the placeholders directly from the text!\\
    
    \# \textbf{Query}
    {}\\
    \{\}\\\hline
    \end{tabular}
    \caption{Natural Language (NL$_1$): 1-shot False Dilemma prompt for LLM experiment}
    \label{tab:5_shot_false_dilemma_NL1}
\end{table*}

\begin{table*}[t]
    \small
    \begin{tabular}{p{170mm}}
    \toprule
   \# \textbf{Task}\\
    Identify the underlying structure of an argument of False Dilemma.\\
    Given a list of fallacy templates, your task is to choose a template that best describes the underlying fallacy structure, choosing the template's placeholders, $[A]$ and $[C]$, directly from the input text. Additionally, the text must be a consecutive sequence of one or more terms without any conjugation. \\
    Please follow the output format.\\
    
    \# \textbf{Definition}s\\
    Entity: a noun phrase in the input.\\
    Event: a verb phrase in the input.\\
    Placeholder: A fill-in-the-blank choice within a template. Each placeholder may either be an entity or an event.\\
    Please note! Placeholders can ONLY be either an entity (i.e., noun phrase) or an event (i.e., verb phrase) and may not be any other type of phrase (e.g., prepositional phrase).\\
    
   \# \textbf{List of Templates}\\
    Template No.1:\\
    Premise 1: An entity/event [A] promotes a good entity/event [C].\\
    Premise 2: An entity/event [¬A] suppresses a good entity/event [C].\\
    Conclusion: Therefore, both Premise 1 and Premise 2 support that [A] should be brought about.\\

    Template No.2:\\
    Premise 1: An entity/event [A] suppresses a bad entity/event [C]\\
    Premise 2: An entity/event [¬A] promotes a bad entity/event [C].\\
    Conclusion: Therefore, both Premise 1 and Premise 2 support that [A] should be brought about.\\

    Template No.3:\\
    Premise 1: An entity/event [A] suppresses a good entity/event [C]\\
    Premise 2: An entity/event [¬A] promotes a good entity/event [C].\\
    Conclusion: Therefore, both Premise 1 and Premise 2 support that [A] should not be brought about.\\

    Template No.4:\\
    Premise 1: An entity/event [A] promotes a bad entity/event [C]\\
    Premise 2: An entity/event [¬A] suppresses a bad entity/event [C].\\
    Conclusion: Therefore, both Premise 1 and Premise 2 support that [A] should not be brought about.\\

    Template No.5:\\
    There is either no consequence in the argument, or the argument cannot be instantiated with one of the templates above.\\

    \# \textbf{Output Format}\\
    Template No.=[No.]\\
    $[A]$=\\
    $[C]$=\\

    \# \textbf{Important Criteria}: Prioritizing entities over events for placeholder.\\
    For choosing placeholder, please prioritize entities over events in the case that an entity itself captures the underlying intent of the argument opposed to the event. However, if the event makes more sense, please choose an event for the placeholder.\\

    \# \textbf{Correct Example}\\
    Input: To get better schools, we have to raise taxes. If we don't, we can't have better schools.\\
    Output:\\
    Template No.=1\\
    $[A]$=raise taxes\\
    $[C]$=schools\\
    Explanation:\\
    Here, there are 2 possible options for [C] which are "schools" (i.e., entity) and "can't have better schools" (i.e., event). Since the entity is the top priority and the second option does not work with template 1 because it is a suppressed relation, "schools" is cchosen for [C].\\
    Also, [A] and [C] are taken directly from the input text. For example, "raising taxes" as [A] also sounds correct, but the term "raising" is not mentioned in the input text. That is why "raise taxes" is chosen for [A]. Because the argument believes that "raise taxes" promote "schools" while not "raise taxes" suppress "school". he conclusion is implicit that “Premise 1 supports that raise taxes should be brought about.” Thus, Template No.=1 is selected.\\
    
    \# \textbf{Wrong Example}\\
    Input: To get better schools, we have to raise taxes. If we don't, we can't have better schools.\\
    Output:\\
    Template No.=1\\
    $[A]$=raising taxes\\
    $[C]$=can't have better schools\\
    Explanation:\\
    Here, there are 2 possible options for [C] which are "schools" (i.e., entity) and "can't have better schools" (i.e., event). However, "can't have better schools" as [C] is incorrect because it is an event instead of the entity of "schools" which already makes sense.\\
    Also, "raising taxes" as [A] is incorrect because the placeholder is not taken directly from the text. Here "raising taxes" is chosen as [A] but the word "raising" does not appear in the input text. Therefore the correct choice for [A] is "raise taxes". \\

    \# Example1\\
    If you can't prove that Ken had an affair with the nanny, then he's been faithful to his wife.\\
    Template No.=1\\
    $[A]$=prove that Ken had an affair with the nanny\\
    $[C]$=he's been faithful to his wife\\

    Again, please only select the placeholders directly from the text!\\

    \# \textbf{Query}\\
    \{\}\\\hline
    \end{tabular}
    \caption{Propositional Logic (PL): 1-shot False Dilemma prompt for LLM experiments.}
    \label{tab:5_shot_false_dilemma_PL}
\end{table*}

\begin{table*}[t]
    \small
    \begin{tabular}{p{170mm}}
    \toprule
    \# \textbf{Task}\\ 
    Identify the underlying structure of an argument of False Dilemma.\\
    Given a list of fallacy templates, your task is to choose a template that best describes the underlying fallacy structure, choosing the template's placeholders, $[A]$ and $[C]$, directly from the input text. Additionally, the text must be a consecutive sequence of one or more terms without any conjugation. \\
    Please follow the output format.\\
    
    \# \textbf{Definition}s\\
    Entity: a noun phrase in the input.\\
    Event: a verb phrase in the input.\\
    Placeholder: A fill-in-the-blank choice within a template. Each placeholder may either be an entity or an event.\\
    Please note! Placeholders can ONLY be either an entity (i.e., noun phrase) or an event (i.e., verb phrase) and may not be any other type of phrase (e.g., prepositional phrase).\\
    
    \# \textbf{List of Templates}\\
    Template No.1:\\
    Premise 1: An entity/event [A] promotes a good entity/event [C].\\
    Premise 2: The absence of an entity/event [A] suppresses a good entity/event [C].\\
    Conclusion: Therefore, [A] should be brought about.\\

    Template No.2:\\
    Premise 1: An entity/event [A] suppresses a bad entity/event [C]\\
    Premise 2: The absence of an entity/event [A] promotes a bad entity/event [C].\\
    Conclusion: Therefore, [A] should be brought about.\\

    Template No.3:\\
    Premise 1: An entity/event [A] suppresses a good entity/event [C]\\
    Premise 2: The absence of an entity/event [A] promotes a good entity/event [C].\\
    Conclusion: Therefore, [A] should not be brought about.\\

    Template No.4:\\
    Premise 1: An entity/event [A] promotes a bad entity/event [C]\\
    Premise 2: The absence of an entity/event [A] suppresses a bad entity/event [C].\\
    Conclusion: Therefore, [A] should not be brought about.\\

    Template No.5:\\
    There is either no consequence in the argument, or the argument cannot be instantiated with one of the templates above.\\

    \# \textbf{Output Format}\\
    Template No.=[No.]\\
    $[A]$=\\
    $[C]$=\\

    \# \textbf{Important Criteria}: Prioritizing entities over events for placeholder.\\
    For choosing placeholder, please prioritize entities over events in the case that an entity itself captures the underlying intent of the argument opposed to the event. However, if the event makes more sense, please choose an event for the placeholder.\\

    \# \textbf{Correct Example}\\
    Input: To get better schools, we have to raise taxes. If we don't, we can't have better schools.\\
    Output:\\
    Template No.=1\\
    $[A]$=raise taxes\\
    $[C]$=schools\\
    Explanation:\\
    Here, there are 2 possible options for [C] which are "schools" (i.e., entity) and "can't have better schools" (i.e., event). Since the entity is the top priority and the second option does not work with template 1 because it is a suppressed relation, "schools" is cchosen for [C].\\
    Also, [A] and [C] are taken directly from the input text. For example, "raising taxes" as [A] also sounds correct, but the term "raising" is not mentioned in the input text. That is why "raise taxes" is chosen for [A]. Because the argument believes that "raise taxes" promote "schools" while not "raise taxes" suppress "school". he conclusion is implicit that “Premise 1 supports that raise taxes should be brought about.” Thus, Template No.=1 is selected.\\
    
    \# \textbf{Wrong Example}\\
    Input: To get better schools, we have to raise taxes. If we don't, we can't have better schools.\\
    Output:\\
    Template No.=1\\
    $[A]$=raising taxes\\
    $[C]$=can't have better schools\\
    Explanation:\\
    Here, there are 2 possible options for [C] which are "schools" (i.e., entity) and "can't have better schools" (i.e., event). However, "can't have better schools" as [C] is incorrect because it is an event instead of the entity of "schools" which already makes sense.\\
    Also, "raising taxes" as [A] is incorrect because the placeholder is not taken directly from the text. Here "raising taxes" is chosen as [A] but the word "raising" does not appear in the input text. Therefore the correct choice for [A] is "raise taxes". \\  

    \#Example1\\
    If you can't prove that Ken had an affair with the nanny, then he's been faithful to his wife.\\
    Template No.=1\\
    $[A]$=prove that Ken had an affair with the nanny\\
    $[C]$=he's been faithful to his wife\\

    Again, please only select the placeholders directly from the text!\\

    \# \textbf{Query}\\
    \{\}\\\hline
    \end{tabular}
    \caption{Natural Language$_2$ (NL$_2$): 1-shot False Dilemma prompt for LLM experiments.}
    \label{tab:5_shot_false_dilemma_NL2}
\end{table*}

\end{document}